\documentclass[letterpaper, 10 pt, conference]{ieeeconf}  
\usepackage{graphicx}
\usepackage{booktabs} 
\usepackage{multirow} 
\usepackage{bbm} 
\usepackage{amssymb}
\usepackage{wrapfig} 
\usepackage{array}
\usepackage{graphicx}
\usepackage{subcaption}
\usepackage{marvosym}
\usepackage[marginal]{footmisc}

\usepackage{bm}
\usepackage{amsmath}
\usepackage{hyperref}
\usepackage{grffile}
\usepackage{pifont} 
\usepackage{mathtools} 
\usepackage{tabularx}
\usepackage{float}
\usepackage{booktabs} 
\usepackage{pifont} 

\usepackage{color}

\usepackage[table]{xcolor}

\IEEEoverridecommandlockouts                              

\title{\LARGE \bf
ASI-Seg: Audio-Driven Surgical Instrument Segmentation with\\ Surgeon Intention Understanding
}

\author{Zhen Chen$^{1,\dag}$, Zongming Zhang$^{1,\dag}$,  Wenwu Guo$^{1}$, Xingjian Luo$^{1}$, Long Bai$^{3}$, Jinlin Wu$^{1,2,*}$, \\Hongliang Ren$^{3}$, Hongbin Liu$^{1,2}$
\thanks{\,\,\,\,This work was supported by the  National Natural Science Foundation of China (Grant No.\#62306313), Hong Kong RGC GRF 14211420, 14203323, NSFC/RGC Joint Research Scheme N\_CUHK420/22, and InnoHK Funding.}
\thanks{\,\,\,\,$^{1}$Z. Chen, Z. Zhang, W. Guo, X. Luo, J. Wu and H. Liu are with Centre for Artificial Intelligence and Robotics (CAIR),
Hong Kong Institute of Science \& Innovation, Chinese Academy of Sciences, Hong Kong SAR, China.}%
\thanks{\,\,\,\,$^{2}$J. Wu and H. Liu are also with State Key Laboratory of Multimodal Artificial Intelligence Systems (MAIS), Institute of Automation, Chinese Academy of Sciences, Beijing, China.}%
\thanks{\,\,\,\,$^{3}$L. Bai and H. Ren are with the Department of Electronic Engineering, Chinese University of Hong Kong, Hong Kong SAR, China.}%
\thanks{\,\,\,\,$\dag$\,Equal contribution, $*$\,Corresponding author.}
}

\begin{document}

\maketitle
\thispagestyle{empty}
\pagestyle{empty}

\begin{abstract}

Surgical instrument segmentation is crucial in surgical scene understanding, thereby facilitating surgical safety. Existing algorithms directly detected all instruments of pre-defined categories in the input image, lacking the capability to segment specific instruments according to the surgeon's intention. During different stages of surgery, surgeons exhibit varying preferences and focus toward different surgical instruments. Therefore, an instrument segmentation algorithm that adheres to the surgeon's intention can minimize distractions from irrelevant instruments and assist surgeons to a great extent. The recent Segment Anything Model (SAM) reveals the capability to segment objects following prompts, but the manual annotations for prompts are impractical during the surgery. To address these limitations in operating rooms, we propose an audio-driven surgical instrument segmentation framework, named ASI-Seg, to accurately segment the required surgical instruments by parsing the audio commands of surgeons. Specifically, we propose an intention-oriented multimodal fusion to interpret the segmentation intention from audio commands and retrieve relevant instrument details to facilitate segmentation. Moreover, to guide our ASI-Seg segment of the required surgical instruments, we devise a contrastive learning prompt encoder to effectively distinguish the required instruments from the irrelevant ones. Therefore, our ASI-Seg promotes the workflow in the operating rooms, thereby providing targeted support and reducing the cognitive load on surgeons. Extensive experiments are performed to validate the ASI-Seg framework, which reveals remarkable advantages over classical state-of-the-art and medical SAMs in both semantic segmentation and intention-oriented segmentation. The source code is available at \href{https://github.com/Zonmgin-Zhang/ASI-Seg}{https://github.com/Zonmgin-Zhang/ASI-Seg}.

\end{abstract}

\section{INTRODUCTION}
Developing computer-assisted surgery systems can improve the quality of interventional healthcare for patients \cite{chen2023surgical,luo2023surgplan,chen2023temporal,bai2024endouic,xu2024transforming}. In particular, surgical instrument segmentation stands as a cornerstone for surgical scene understanding \cite{Padoy_unsupervised,ISINet,yue_2019,sun2023pwiseg}, which can benefit visual navigation, precise operation, and instrument tracking, thereby facilitating surgical safety and patient outcomes.

To achieve accurate instrument segmentation, existing works \cite{TernausNet,ISINet,Zhao_2020,Baby2022FromFT,rethinking_wangan} have conducted a lot of research from different aspects. For instance, the ISINet \cite{ISINet} further improved the TernausNet \cite{TernausNet} by identifying instrument candidates and assigning category labels. In addition, the Dual-MF \cite{Zhao_2020} utilized the motion flow of surgical instruments to benefit segmentation, and the S3Net \cite{Baby2022FromFT} focused on discriminating instrument categories. Despite great progress, these works directly segmented all instruments of pre-defined categories in the input image, lacking the capability to segment specific instruments according to the surgeon's intention. In clinical practice, surgeons exhibit varying preferences and focus toward different surgical instruments during various stages of the surgery. Therefore, surgical instrument segmentation algorithms that adhere to the surgeon's intention are highly demanded.

The recent advent of the segment anything model (SAM) \cite{kirillov2023_sam} has revealed the superior robustness and adaptability in natural images in various scenarios. On this basis, SAM has begun to penetrate into the field of medical imaging and demonstrated its capabilities in medical image segmentation \cite{cheng2023sam_med2d,zhang2024segment_sam_survery,ma2024sam_majun_nc,huang2024segment}. In particular, SAM can segment specific objects based on manual prompts, showing the possibility of segmenting surgical instruments on demand in the operating room. But most existing medical SAM studies \cite{ma2024sam_majun_nc,lin2023samus} rely on more manual annotations, by labeling points or bounding boxes as the prompt. The extensive use of manual annotations interrupts surgical workflows, which is impractical in the operating rooms. Therefore, the ideal surgical instrument segmentation algorithm should eliminate the need for manual annotation, and automatically segment the required surgical instruments based on the intention of surgeons.

To address these limitations of surgical instrument segmentation in the operating rooms, we propose an audio-driven surgical instrument segmentation framework, named ASI-Seg, to accurately segment the required surgical instruments by parsing the audio commands of surgeons. Specifically, we propose an intention-oriented multimodal fusion to interpret the segmentation intention from audio commands, and retrieve relevant instrument details to facilitate segmentation. Moreover, to guide our ASI-Seg segment of the required surgical instruments, we devise a contrastive learning prompt encoder to effectively distinguish the required instruments from the irrelevant ones. In this way, our ASI-Seg promotes the workflow in the operating rooms, thereby providing targeted support and reducing the cognitive load on surgeons. 

The contributions of this work are summarized as follows:
\begin{itemize}  
\item We propose the ASI-Seg framework to achieve audio-driven surgical instrument segmentation based on the surgeon's intention.

\item We devise an {intention-oriented multimodal fusion} to interpret the intention and retrieve details for ASI-Seg.

\item We devise a {contrastive learning prompt encoder} to distinguish the required instruments from irrelevant ones.

\item Extensive experiments on the EndoVis2018 and EndoVis2017 datasets confirm the superior performance of ASI-Seg in both semantic segmentation and intention-oriented segmentation.

\end{itemize}

\section{Related work}

\noindent\textbf{Surgical Instrument Segmentation.}
Existing works \cite{TernausNet,ISINet,Zhao_2020,Baby2022FromFT,rethinking_wangan} conducted surgical instrument segmentation from different aspects. In particular, the TernausNet \cite{TernausNet} improved the network structure to achieve accurate instrument segmentation. The ISINet \cite{ISINet} achieved the segmentation by identifying instrument candidates and assigning category labels. The Dual-MF \cite{Zhao_2020} utilized the motion flow of surgical instruments for more accurate segmentation decoding. The S3Net \cite{Baby2022FromFT} further addressed the difficulty in discriminating instrument categories. In addition, Wang \textit{et al.} \cite{rethinking_wangan} blended the irrelevant tissues with required instruments to facilitate segmentation with augmented samples. Different from these works that directly segmented all instruments of pre-defined categories, our ASI-Seg can segment specific instruments according to the surgeon's intention.

\noindent\textbf{The SAM for Medical Imaging.}
By leveraging both sparse (\textit{e.g.}, point, box, and text) and dense (\textit{e.g.}, mask) prompts, the segment anything model (SAM) \cite{kirillov2023_sam} has well revealed the advantage in image segmentation across a variety of scenarios. To transfer SAM to downstream scenarios, existing works adopted different fine-tuning strategies, including directly fine-tuning the image encoder \cite{shui2023unleashing_sam_yanglin} or mask decoder \cite{ma2024sam_majun_nc}, and using the parameter efficient fine-tuning (\textit{e.g.}, the low-rank adaptation (LoRA) \cite{hu2022lora} and adapter \cite{houlsby2019parameter}), considering the huge amount of SAM parameters. Many medical SAM works \cite{ma2024sam_majun_nc,huang2024segment,lin2023samus,zhang2023customized_sam_liudong,wu2023medsa_jinyueming,chen2024sam} have been investigated to customize segmentation capability to medical imaging. Huang \textit{et al.} \cite{huang2024segment} explored the impact of different prompts on medical image segmentation, and the MedSAM \cite{ma2024sam_majun_nc} further fine-tuned the SAM with bounding box prompts on large-scale medical image datasets. For the surgical images, the SurgicalSAM \cite{jieboluo2024surgicalsam} introduced the class prototypes and designated target class to guide the segmentation with the category information. In general, most medical SAMs either demand huge computational resources in fine-tuning \cite{shui2023unleashing_sam_yanglin} or rely on manual annotations for prompt during inference \cite{ma2024sam_majun_nc,huang2024segment,lin2023samus,zhang2023customized_sam_liudong,wu2023medsa_jinyueming}, which is impractical for clinical usage.

\section{Methodology}
\subsection{Overview of ASI-Seg Framework}\label{sec_overview}
\begin{figure*}[tb]
  \centering
  \includegraphics[width=0.99\linewidth]{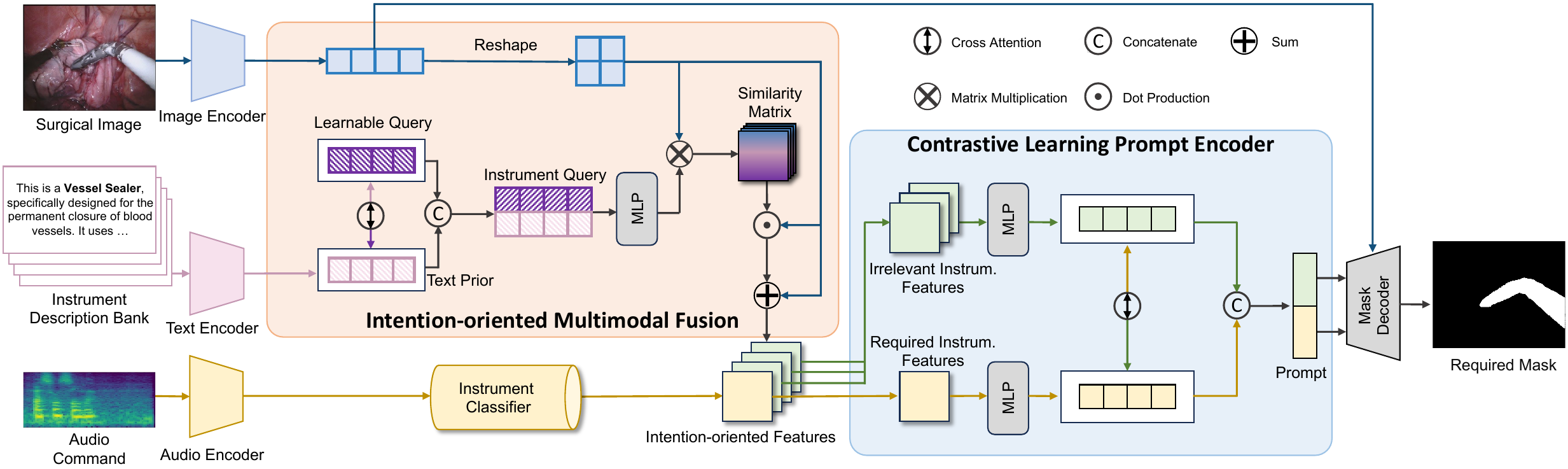} 
  \caption{\textbf{Overview of ASI-Seg.} The ASI-Seg mainly consists of the Intention Multimodal Fusion and the Contrastive Learning Prompt Encoder. By parsing the segmentation intention of surgeons, the ASI-Seg first exploits the multimodal knowledge to generate the intention-oriented features and then performs the contrastive learning between required instrument features and the irrelevant ones to produce prompt 
     for segmenting the required instruments.}
  \label{fig:subfig1}
\end{figure*}

In this work, we propose the ASI-Seg framework to segment required surgical instruments by following the audio commands of surgeons. As elaborated in Fig.~\ref{fig:subfig1}, we propose the ASI-Seg framework to segment required surgical instruments by following the audio commands of surgeons. Given a surgical image, the ASI-Seg pareses the audio command for segmentation intention and generates the required instrument masks to meet the demand of surgeons.

\subsection{Intention-Oriented Multimodal Fusion}\label{sec_scl}
To obtain the features of the surgeon's specified instruments, we propose an Intention-Oriented Multimodal Fusion module in this section. Firstly, we propose an audio intention recognition module to predict the surgeon's segment intention. Then, we propose a text fusion module and a visual fusion module to inject detailed language description information and richer visual information into a group of learnable queries. Lastly, we utilize the recognized audio intention to select the intention-oriented features.

\noindent\textbf{Audio Intention Recognition.} We sample the discretion audio signals $a'$ from raw audio signals $a$ with 16K Hz. Then, we transfer the discretion audio signals to the Mel spectrogram as follows:
\begin{equation}
    A_{\text{mel}} = \pi(a, a', C_s, W_s, s), \\
\end{equation}
where $\pi$ is the Mel spectrogram transformation \cite{librosa}, $C_s$ is the channel size, $W_s$ is the window size and $s$ is the stride size. For better numerical calculations, we further normalize the scale of $A_{\text{mel}}$ to the range of $[-1, 1]$, as follows:
\begin{equation}
    A_{\text{norm}} = \frac{2* (A_{\text{mel}} - \mu)}{\max(A_{\text{mel}}) - \min(A_{\text{mel}})} - 1,
\end{equation}
where $\mu$ is the mean of Mel spectrogram $A_{\text{mel}}$ among the training data. To predict the intention of surgeons, we feed  $A_{\text{norm}}$ to an Audio Encoder $E_A$ and an audio classifier $\phi$:
\begin{equation}
    \mathcal{C} = \phi(E_A(A_{\text{norm}})),
\end{equation}
where $\mathcal{C}$ is the audio intention recognition result.

\noindent\textbf{Text Fusion.}
The surgeon's audio commands may only include the names of instruments. These high-level commands make it challenging for the model to capture the necessary features of the required instruments from visual information. Therefore, we incorporate detailed textual descriptions of each instrument into the learnable query as a complement to the high-level audio commands. 

Specifically, we first use a Text Encoder $E_T$ to extract textual features $f_t$ from a pre-prepared Instrument Description Bank $\{B_k\}^{K}_{k=1}$, which stores detailed descriptions of $K$ instruments as follows:
\begin{equation}
    f_t = \text{concat}(E_T(\{B_i\}^{K}_{k=1})),
\end{equation}
where $B_k$ refers to the specific instrument description of $k$-th instrument, and $f_t\in \mathbb{R}^{K \times d}$ is the concatenated textual feature of all $K$ instruments with feature dimension $d$.

Then, we initialize the $K$ learnable queries $f_c$ corresponding to each surgical instrument.
These queries $f_c$ are then fused with textual feature $f_t$ through a mutual cross-attention module to form an Instrument Query $q$, as follows: 
\begin{equation}
\begin{split}
        q_t &= \text{softmax}(\frac{Q_tK_c^T}{\sqrt{D}})V_c, \\
        q_c &= \text{softmax}(\frac{Q_cK_t^T}{\sqrt{D}})V_t, \\
        q &= \text{MLP}(\text{concat}(q_t,q_c)),
\end{split}
\end{equation}
where $Q_t$, $Q_c$, $K_t$, $K_c$, $V_t$, $V_c$ are attention queries, keys and values from textual feature $f_t$ and learnable query $f_c$ correspondingly. $D$ is the dimension of the keys and values. The instrument query $q\in \mathbb{R}^{K \times d}$ is a fused result of $q_t \in \mathbb{R}^{K \times d}$ and $q_c\in \mathbb{R}^{K\times d}$. We concatenate them and use a MLP to reduce the concatenated dimension $2d$ to $d$. These instrument queries fused with detailed textual knowledge can provide more distinguishable information between surgical instruments.

\noindent\textbf{Visual Fusion.} We leverage the instrument queries to extract cross-modality visual information from the input image. First, we extract and reshape the image features $f_i \in \mathbb{R}^{h \times w \times d}$ from the image $I \in \mathbb{R}^{H \times W \times 3}$ by an Image Encoder $E_I$ 
as follows:
\begin{equation}
    f_i = E_I(I),
\end{equation}
where $H$, $W$ are the original image size and $h$, $w$ are the reshaped size, $d$ is the feature dimension. Then, we compute the similarity between Instrument Query $q$ and image feature $f_i$ to get a sequence of Similarity Matrix $ \{S^k|S^k=q^k \cdot f_i\}^{K}_{k=1}$, where $q_k$ and $S^k$ are the corresponding query and Similarity Matrix of instrument $n$. Finally, we add the image feature $f_i$ to the Similarity Matrix sequence $\{S^k\}^{K}_{k=1}$ to get the Multimodal Features $F =  \{f_{i\text{-}t}^k\}^K_{k=1}$ that contains both image and textual information as follows:  
\begin{equation}
    \{f_{i\text{-}t}^n\}^K_{k=1} = \{f_i \cdot S^k + f_i\}^K_{k=1},
\end{equation}
where $f_i$ represents the image feature and $S^k$ refers to the Similarity Matrix of instrument $k$. 

\noindent\textbf{Feature Assignment with Audio Intention.} We use $\mathcal{C}$ to divide the Multimodal Feature $F$ into required feature sequence $F^{+}$
and irrelevant feature sequence $F^{-}$, as follows: 
\begin{equation}
\begin{split}
    F^{+} &= \{f_{i\text{-}t}^\mathcal{C}\}, \\
    F^{-} &= \{f_{i\text{-}t}^{k, k \neq \mathcal{C}}\}^{K}_{k=1},
\end{split}
\end{equation}
where $F^{+}$ refers to feature of the current segment target $\mathcal{C}$, $F^{-}$ refers to the rest of the features in $F$, and $f_{i\text{-}t}^k$ is the multimodal feature of surgical instrument $k$. Following the surgeon's intention, this approach divides the multimodal features $F$ into two groups of Intention-oriented Features.

\subsection{Contrastive Learning Prompt Encoder}\label{sec_scl}
To effectively distinguish between required and irrelevant instrument features, we design the Contrastive Learning Prompt Encoder to provide the mask decoder with the specific prompt of the instrument to be segmented.

\noindent\textbf{Distinguishing Cross-Attention.} We employ a mutual cross-focusing mechanism between the required instrument feature $F^{+}$ and the irrelevant instrument feature $F^{-}$, which aims to enhance the focus on the unique properties of the surgical instruments to be segmented. Firstly, we compute the attention similarity to obtain easily confounded regions as follows:
\begin{equation}
\text{Attention}(F^{+}, F^{-}) = \text{softmax}\left(\frac{Q_{F^{+}}K_{F^{-}}^T}{\sqrt{D}}\right)V_{F^{-}},
\end{equation}
where $Q_{F^{+}}$, $K_{F^{-}}$, $V_{F^{-}}$are attention query, key, and value from the required instrument feature $F^{+}$ and the irrelevant instrument feature $F^{-}$ correspondingly. In addition,  $\text{Attention}(F^{-}, F^{+})$ is the same.

Then, we adopt an inverse residual mechanism as follows:
\begin{equation}
P^* = P - \text{Attention}(F^{+}, F^{-}),
\end{equation}
where $P^*$ is the output required instrument feature. $P^*$ eliminates information similar to the irrelevant instrument feature and maintains its unique attributes and characteristics, which is essential for accurate segmentation.

\noindent\textbf{Contrastive  Learning.}
To further push relevant instrument features and irrelevant instrument features to be separable, we design a contrast learning between the required instrument features and the irrelevant ones. Specifically, the contrastive learning loss $\mathcal{L}_{\rm CL}$ is defined using three parameters, including the required instrument features $P$, the irrelevant instrument features $N$, as well as features $v$ from image embeddings filtered through ground truth masks. The formula is as follows:
\begin{equation}
\mathcal{L}_{\rm CL} = -\frac{1}{K} \sum_{n=1}^K \log \frac{\exp(P^{(\mathcal{C})} \cdot v^{(\mathcal{C})} / \tau)}{\sum_{n=1}^K \exp(P^{(\mathcal{C})} \cdot v^{(n)} / \tau)},
\end{equation}
where $\tau$ refers to the temperature factor, and $P^{(\mathcal{C})}$ represents the required instrument features of class $\mathcal{C}$. This contrastive loss pushes the required instrument away from the irrelevant instrument features, enhancing the feature discrimination capability of our ASI-Seg.

\noindent\textbf{Mask Decoder.} We adapt the SAM \cite{kirillov2023_sam} mask decoder to generate the mask of the required instruments. Our ASI-Seg exhibits enhanced differentiation between required instruments and irrelevant instruments using tailored contrastive learning. This distinction significantly augments the segmentation capability of the SAM mask decoder.  We regard the features derived from required instruments as foreground prompts and the features obtained from irrelevant instruments as background prompts for the SAM mask decoder. Therefore, the ASI-Seg can generate the accurate mask of the required surgical instruments with comprehensive prompts.

\begin{table}[t]
\centering
\caption{Semantic Segmentation Comparison on the EndoVis2018 Dataset.}\label{tab:tab1}
\begin{tabular}{l|cc}
\toprule
\textbf{Method} & \textbf{Challenge IoU} & \textbf{IoU} \\
\midrule
TernausNet \cite{TernausNet} & 46.22 & 39.87 \\
MF-TAPNet \cite{yue_2019} & 67.87 & 39.14 \\
Dual-MF \cite{Zhao_2020} & 70.40 & - \\
ISINet \cite{ISINet} & 73.03 & 70.94 \\
S3Net \cite{Baby2022FromFT} & 75.81 & 74.02 \\
MaskTrack-RCNN \cite{Yang_2019_ICCV} + SAM \cite{kirillov2023_sam} & 78.49 & 78.49 \\
Mask2Former \cite{Cheng_2022_CVPR} + SAM \cite{kirillov2023_sam} & 78.72 & 78.72 \\
TrackAnything (1 Point) \cite{Yang2023TrackAS} & 40.36 & 38.38 \\
TrackAnything (5 Points) \cite{Yang2023TrackAS} & 65.72 & 60.88 \\
PerSAM \cite{zhang2024personalize} & 49.21 & 49.21 \\
PerSAM (Fine-Tune) \cite{zhang2024personalize} & 52.21 & 52.21 \\
SurgicalSAM \cite{jieboluo2024surgicalsam} & 80.33 & 80.33 \\
\rowcolor[rgb]{ .949,  .949,  .949} ASI-Seg (Ours) & \textbf{82.37} & \textbf{82.37} \\
\bottomrule
\end{tabular}
\end{table}

\subsection{Optimization}
In the training of ASI-Seg, we freeze the image encoder, the audio encoder and the text encoder with massive parameters, and merely optimize the lightweight instrument classifier and mask decoder, as well as the proposed intention-oriented multimodal fusion and contrastive learning prompt encoder, which makes the end-to-end training efficient. The ASI-Seg is optimized by two loss terms, as follows:
\begin{equation}
\mathcal{L}=\mathcal{L}_{\rm DICE}+\mathcal{L}_{\rm CL},
\end{equation}
where the dice loss $\mathcal{L}_{\rm DICE}$ \cite{V-Net} is for segmentation and the contrastive learning loss $\mathcal{L}_{\rm CL}$ is used to dynamically update the learnable query in the ASI-Seg. In this way, ASI-Seg is capable of segmenting the required instruments according to the intention of surgeons.

\section{Experiment}
\subsection{Implementation and Datasets} 
\noindent\textbf{Dataset.} We perform the comprehensive evaluation on the EndoVis2018 \cite{Allan2018Endovis} and EndoVis2017 datasets \cite{Allan2017Endovis}. To guarantee fair comparisons, we follow the standard protocol \cite{ISINet,TernausNet}. Specifically, the EndoVis2017 dataset, comprising eight videos, is subjected to a 4-fold cross validation \cite{TernausNet}. The video sequences with a high resolution of $1,280\times 1,024$ are acquired
from \textit{da Vinci Xi} surgical system during different porcine procedures. Meanwhile, the EndoVis2018 dataset encompasses 11 training videos alongside four validation videos, thereby presenting a comprehensive platform for benchmarking. Both datasets feature seven unique categories of surgical instruments, enabling an in-depth assessment of our segmentation effectiveness.

\begin{table}[t]
\setlength{\tabcolsep}{3.5mm}
\centering
\caption{Semantic Segmentation Comparison on the EndoVis2017 Dataset.}\label{tab:tab3}
\begin{tabular}{l|cc}
\toprule
\textbf{Method} & \textbf{Challenge IoU} & \textbf{IoU} \\
\midrule
TernausNet \cite{TernausNet} & 35.27 & 12.67 \\
MF-TAPNet \cite{yue_2019} & 37.25 & 13.49 \\
Dual-MF \cite{Zhao_2020} & 45.80 & - \\
ISINet \cite{ISINet} & 55.62 & 52.20 \\
TraSeTr \cite{track_Zhao_2022} & 60.40 & - \\
S3Net \cite{Baby2022FromFT} & 72.54 & 71.99 \\
Mask2Former \cite{Cheng_2022_CVPR} + SAM \cite{kirillov2023_sam} & 66.21 & 66.21 \\
TrackAnything (1 Point) \cite{Yang2023TrackAS} & 54.90 & 52.46 \\
TrackAnything (5 Points) \cite{Yang2023TrackAS} & 67.41 & 64.50 \\
PerSAM \cite{zhang2024personalize} & 42.47 & 42.47 \\
PerSAM (Fine-Tune) \cite{zhang2024personalize} & 41.90 & 41.90 \\
SurgicalSAM \cite{jieboluo2024surgicalsam} & 69.94 & 69.94 \\
\rowcolor[rgb]{ .949,  .949,  .949} ASI-Seg (Ours) & \textbf{71.64} & \textbf{71.64} \\
\bottomrule
\end{tabular}
\end{table}

\noindent\textbf{Implementation Details.} 
We implement our ASI-Seg in PyTorch on a single NVIDIA A800 GPU. In our ASI-Seg, we use the pre-trained ViT \cite{dosovitskiy2021an} as the image encoder, and use the text encoder of CLIP \cite{clip} as the text encoder, and the pre-trained audio encoder \cite{whisper} as our audio encoder. Additionally, we randomly initialize audio embeddings as the category query. To enhance architectural stability and concentrate on novel components, we maintain static image encoders while dynamically updating the learnable query and mask decoder weights. We set the temperature factor \(\tau\) of contrastive loss as \(0.07\), Adam as the optimizer with a learning rate of \(0.0001\) across both datasets to accommodate their distinct complexities. The training leverages pre-computed image embeddings and a batch size of \(16\) for EndoVis2017 and \(64\) for EndoVis2018 datasets.

\noindent\textbf{Evaluation Metrics.} We perform the evaluation using three critical segmentation metrics following \cite{jieboluo2024surgicalsam}, including the Challenge IoU \cite{Allan2017Endovis}, IoU, and mean class IoU (mc IoU) \cite{ISINet,Baby2022FromFT}. These metrics ensure our ASI-Seg is rigorously measured and validated against these benchmarks.

\begin{figure*}[tb]
  \centering
  \includegraphics[width=0.97\linewidth]{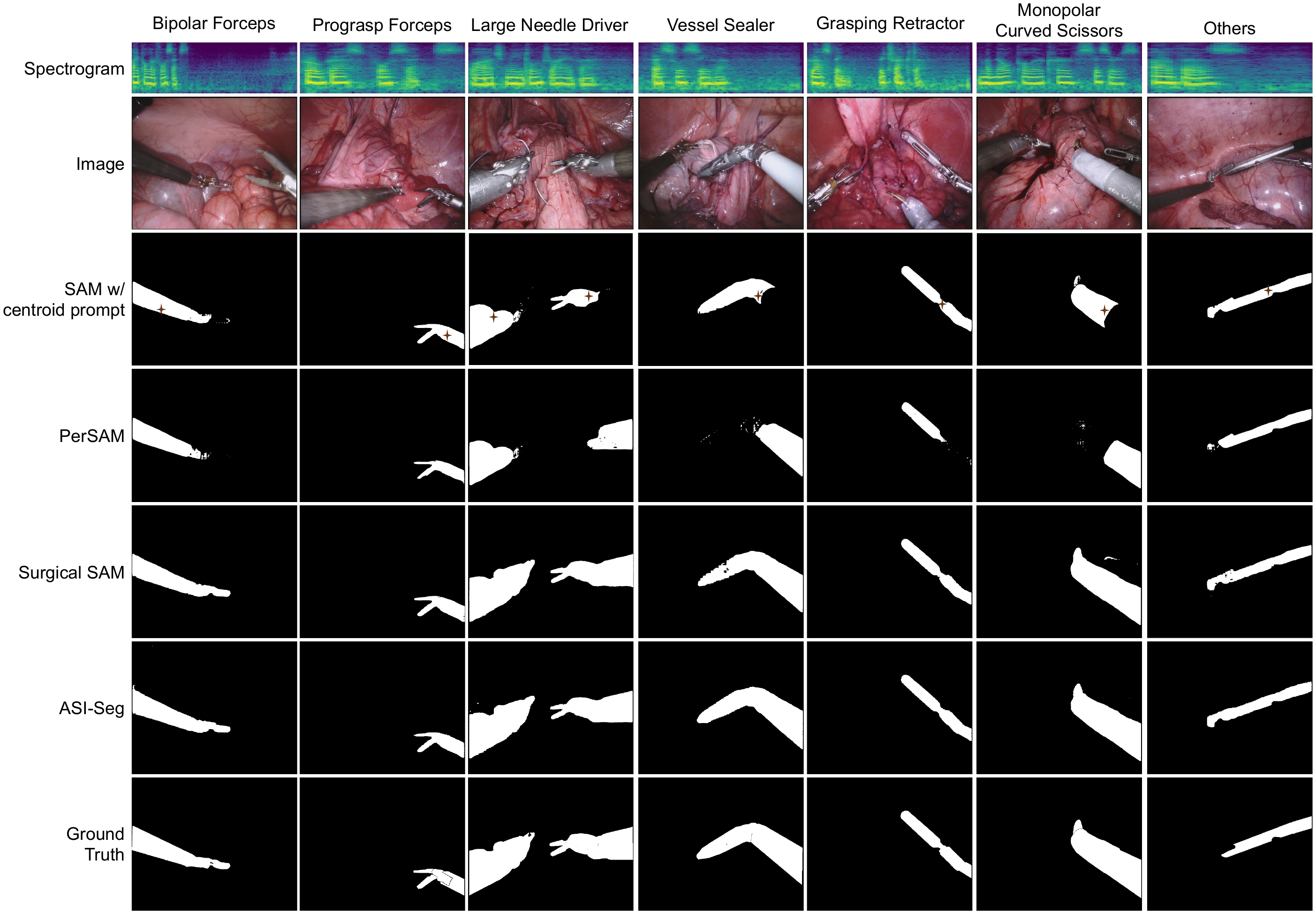} 
  \caption{Qualitative comparison of intention-oriented segmentation on the EndoVis2017 dataset.}
  \label{fig:spectro}
\end{figure*}

\subsection{Comparisons with State-of-the-arts}
We conduct the comprehensive comparison between our ASI-Seg framework and state-of-the-art surgical instrument segmentation methods and advanced SAM approaches on the EndoVis2018 \cite{Allan2018Endovis} and EndoVis2017 \cite{Allan2017Endovis} datasets.

\begin{table*}[!htbp]
\setlength{\tabcolsep}{3.5mm}
\centering
\caption{Intention-oriented Segmentation Comparison on the EndoVis2018 Dataset.}\label{tab:tab2}
\begin{tabular}{l|ccccccccc}
\toprule
\textbf{Method} & \textbf{mc IoU} & \textbf{BF} & \textbf{PF} & \textbf{LND} & \textbf{SI} & \textbf{CA} & \textbf{MCS} & \textbf{UP} \\
\midrule
MaskTrack-RCNN \cite{Yang_2019_ICCV} + SAM \cite{kirillov2023_sam} & 56.07 & 79.83 & 74.86 & 43.12 & 62.88 & 16.74 & 91.62 & 23.45 \\
Mask2Former \cite{Cheng_2022_CVPR} + SAM \cite{kirillov2023_sam} & 52.50 & \textbf{85.95} & \textbf{82.31} & 44.08 & 0.00 & \textbf{49.80} & \textbf{92.17} & 13.18 \\
TrackAnything (1 Point) \cite{Yang2023TrackAS} & 20.62 & 30.20 & 12.87 & 24.46 & 9.17 & 0.19 & 55.03 & 12.41 \\
TrackAnything (5 Points) \cite{Yang2023TrackAS} & 38.60 & 72.90 & 31.07 & \textbf{64.73} & 10.24 & 12.28 & 61.05 & 17.93 \\
PerSAM \cite{zhang2024personalize} & 34.55 & 51.26 & 34.40 & 46.75 & 16.45 & 15.07 & 52.28 & 25.62 \\
PerSAM (Fine-Tune) \cite{zhang2024personalize} & 37.24 & 57.19 & 36.13 & 53.86 & 14.34 & 25.94 & 54.66 & 18.57 \\
SurgicalSAM \cite{jieboluo2024surgicalsam} & 58.87 & 83.66 & 65.63 & 58.75 & 54.48 & 39.78 & 88.56 & 21.23 \\
\rowcolor[rgb]{ .949,  .949,  .949} ASI-Seg (Ours) & \textbf{64.18} & 83.12 & 65.87 & 59.24 & \textbf{90.43} & 34.90 & 60.10 & \textbf{55.62} \\
\bottomrule
\end{tabular}
\end{table*}

\noindent{}\textbf{Semantic Segmentation Analysis.} 
As shown in Table~\ref{tab:tab1} and Table~\ref{tab:tab3}, we first perform the comparison of semantic segmentation by segmenting all the instruments in the input image on the EndoVis2018 and EndoVis2017 datasets, respectively. In general, our ASI-Seg achieves the best performance in the semantic segmentation landscape, with IoU of $82.37\%$ and $71.64\%$ on the EndoVis2018 and EndoVis2017 datasets, respectively.  Note that our ASI-Seg outperforms the second-best method \cite{jieboluo2024surgicalsam} with an IoU advantage of $2.04\%$ and $1.70\%$ on these two datasets. These improvements indicate a profound improvement of our ASI-Seg in the model capacity to distinguish surgical instruments from irrelevant ones and complex backgrounds.

\begin{table*}[t]
\setlength{\tabcolsep}{3.5mm}
\centering
\caption{Intention-oriented Segmentation Comparison on the EndoVis2017 Dataset.}\label{tab:tab4}
\begin{tabular}{l|cccccccc}
\toprule
\textbf{Method} & \textbf{mc IoU} & \textbf{BF} & \textbf{PF} & \textbf{LND} & \textbf{VS} & \textbf{GR} & \textbf{MCS} & \textbf{UP} \\
\midrule
Mask2Former \cite{Cheng_2022_CVPR} + SAM \cite{kirillov2023_sam} & 55.26 & 66.84 & \textbf{55.36} & \textbf{83.29} & 73.52 & 26.24 & 36.26 & 45.34 \\
TrackAnything (1 Point) \cite{Yang2023TrackAS} & 55.35 & 47.59 & 28.71 & 43.27 & 82.75 & \textbf{63.10} & 66.46 & 55.54 \\
TrackAnything (5 Points) \cite{Yang2023TrackAS} & 62.97 & 55.42 & 44.46 & 62.43 & \textbf{83.68} & 62.59 & 67.03 & \textbf{65.17} \\
PerSAM \cite{zhang2024personalize} & 41.80 & 53.99 & 25.89 & 50.17 & 52.87 & 24.24 & 47.33 & 38.16 \\
PerSAM (Fine-Tune) \cite{zhang2024personalize} & 39.78 & 46.21 & 28.22 & 53.12 & 57.98 & 12.76 & 41.19 & 38.99 \\
SurgicalSAM \cite{jieboluo2024surgicalsam} & 67.03 & 68.30 & 51.77 & 75.52 & 68.24 & 57.63 & 86.95 & 60.80 \\
\rowcolor[rgb]{ .949,  .949,  .949} ASI-Seg (Ours) & \textbf{68.37} & \textbf{73.92} & 47.61 & 80.33 & 75.44 & 52.60 & \textbf{89.78} & 58.90 \\
\bottomrule
\end{tabular}
\end{table*}

\noindent{}\textbf{Intention-oriented Segmentation Analysis.} 
To evaluate the capability of the ASI-Seg, we perform the comparison of intention-oriented segmentation, as shown in Table~\ref{tab:tab2} and Table~\ref{tab:tab4} for EndoVis2018 and EndoVis2017 datasets, respectively. This comparison encompasses a broad spectrum of surgical instruments, including Bipolar Forceps (BF), Prograsp Forceps (PF), Large Needle Driver (LND), Suction Instrument (SI), Vessel Sealer (VS), Clip Applier (CA), Grasping Retractor (GR), Monopolar Curved Scissors (MCS), and Ultrasound Probe (UP). Specifically, we calculate the IoU of each category with segmentation intention and average them for the mean class IoU (mc IoU). Our ASI-Seg achieves the superior mc IoU of $64.18\%$ and $68.17\%$ in the EndoVis2018 and EndoVis2017 datasets, with overwhelming improvement over advanced SAM approaches. In particular, our ASI-Seg reveals the advantage of $5.31\%$ in mc IoU over the second-best SurgicalSAM with category prompt \cite{jieboluo2024surgicalsam} on the EndoVis2018 dataset, and also achieves the balanced segmentation performance in different categories of surgical instruments, as shown in Table~\ref{tab:tab2}. As such, these comparisons confirm the versatility of our ASI-Seg across a wide range of surgical scenarios to meet the requirements of surgeons.

Furthermore, we qualitatively compare the segmentation masks of the ASI-Seg and SAM-based approaches with the intention of segmenting instruments of each category, as illustrated in Fig.~\ref{fig:spectro}. It is worth noting that our ASI-Seg does not require manual annotations \cite{kirillov2023_sam} or assigned category \cite{jieboluo2024surgicalsam} for prompt. In the comparison, our ASI-Seg correctly understands the segmentation intention and generates the most accurate masks of the required instruments. These enhancements highlight the proficiency of ASI-Seg in identifying and classifying diverse surgical instruments. In general, the performance advantage of ASI-Seg substantiates the superiority in surgical instrument segmentation.

\begin{figure}[tb]
  \centering
  \includegraphics[width=0.99\linewidth]{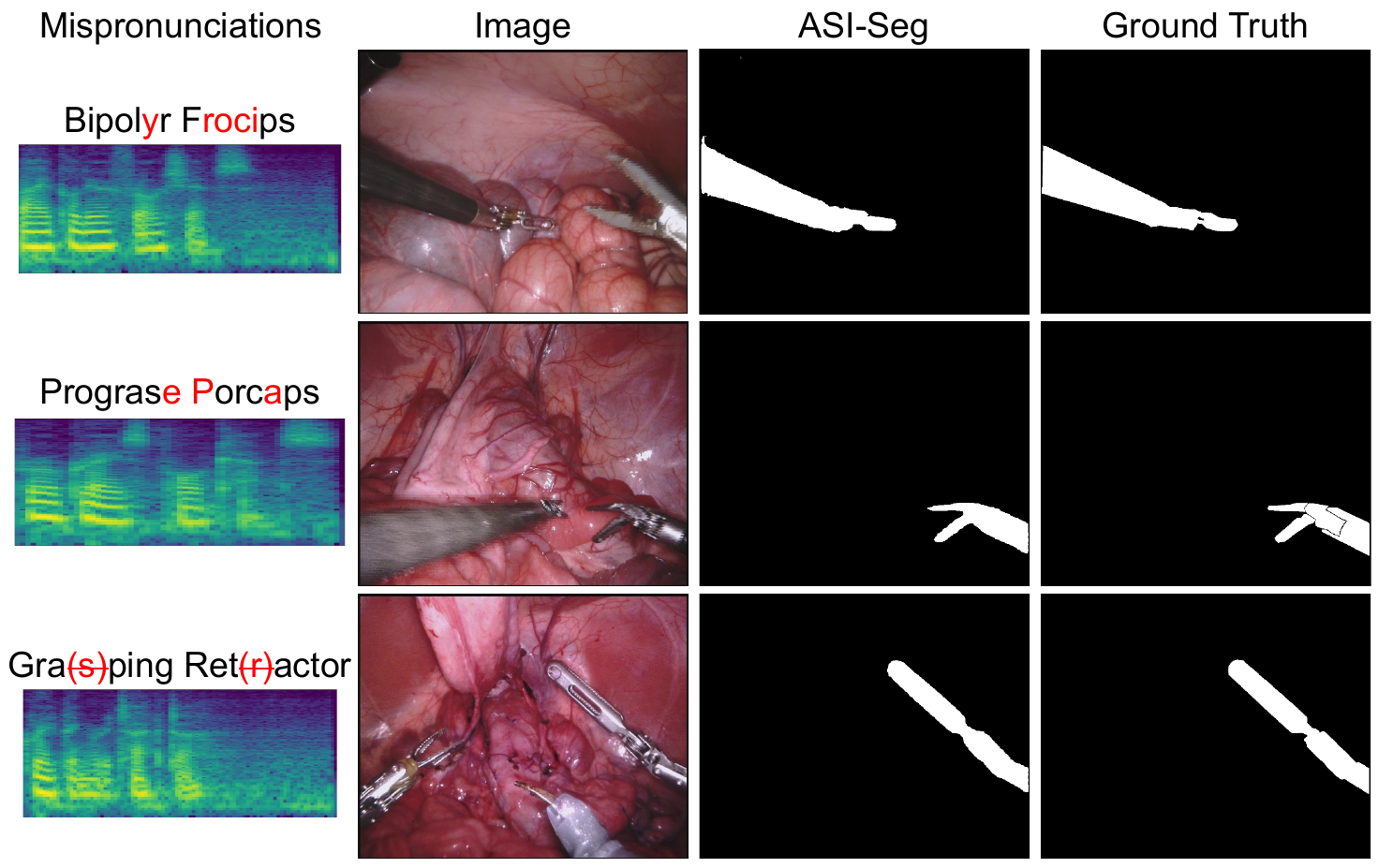} 
  \caption{Impact of mispronunciations on our ASI-Seg.}
  \label{fig:robust}
\end{figure}

\subsection{Robustness Study}
We further investigate the robustness of our ASI-Seg against defective audio commands, \textit{e.g.}, the mispronunciation of instrument names. As illustrated in Fig. \ref{fig:robust}, when there are obvious mispronunciations in the input audio, \textit{e.g.},  surgeons may mistakenly articulate the Bipolar Forceps as \textit{Bipolyr Frocips}, our ASI-Seg is still capable to recognize the intention into the correct instrument category and complete accurate segmentation. These results confirm the robustness of our ASI-Seg to identify instruments that surgeons intend to use despite verbal errors, which is also an advantage compared to text instructions.

\subsection{Ablation Study}

To validate the effectiveness of the proposed modules, we perform the ablation study on the EndoVis2018 \cite{Allan2018Endovis} dataset, as shown in Table~\ref{tab:tab5}. Compared with the vanilla baseline, our framework with the instrument description bank gains a $8.42\%$ increase in mc IoU. This confirms that the integration of textual knowledge is a pivotal component in cultivating distinct learnable queries for different categories, thereby ameliorating the precision of instrument segmentation. On the other hand, our framework obtains a $4.98\%$ increase in mc IoU when adding the contrastive learning in ASI-Seg. In our ASI-Seg, the advantage provided by contrastive learning is mainly attributed to its ability to dynamically emphasize the required instrument features while attenuating irrelevant ones. Therefore, the contrastive learning enables the ASI-Seg to become more proficient in differentiating instruments by focusing on the attributes necessary for differentiation. In this way, the proposed ASI-Seg benefits from these tailored designs, resulting in the performance advantage in surgical instrument segmentation at the operating rooms.

\begin{table}[tbp]
\setlength{\tabcolsep}{3mm}
\centering
\caption{Ablation Study on our ASI-Seg.}\label{tab:tab5}
\scalebox{0.80}{
\begin{tabular} 
{@{} c c | c c c c @{}}
\toprule
{Instrument}  & \multirow{2}{*}{Contrastive Learning} & \multirow{2}{*}{Challenge IoU} & \multirow{2}{*}{IoU} & \multirow{2}{*}{mc IoU} \\
{Description Bank}&&&&\\
\midrule
&  & 76.14 & 76.14 & 51.00 \\
\ding{51} & & 80.17 & 80.17 & 59.42 \\
& \ding{51}  & 78.63 & 78.63 & 55.98 \\
\ding{51} & \ding{51} & \textbf{82.37} & \textbf{82.37} & \textbf{64.18} \\
\bottomrule
\end{tabular}
}
\end{table}

\section{CONCLUSIONS}
In this work, we propose the ASI-Seg framework to accurately segment the required surgical instruments by parsing the audio commands of surgeons. In our ASI-Seg framework, the intention-oriented multimodal fusion can interpret the segmentation intention and retrieve relevant instrument details to facilitate segmentation. Moreover, the contrastive learning prompt encoder can distinguish the required instruments from the irrelevant ones to guide our ASI-Seg segment of the required surgical instruments. Therefore, our ASI-Seg can minimize distractions from irrelevant instruments assist surgeons to a great extent, and promote the workflow in the operating rooms. Extensive experiments are performed to validate the ASI-Seg framework, which reveals remarkable advantages over classical state-of-the-art and advanced SAMs in both semantic segmentation and intention-oriented segmentation.










\bibliographystyle{IEEEtran}
\bibliography{mybibliography}

\end{document}